# Domain-Specific Text Generation for Machine Translation

**Yasmin Moslem**  yasmin.moslem@adaptcentre.ie
**Andy Way**  andy.way@adaptcentre.ie
School of Computing, Dublin City University, Dublin, Ireland
**Rejwanul Haque**  rejwanul.haque@adaptcentre.ie
School of Computing, National College of Ireland, Dublin, Ireland
**John D. Kelleher**  john.kelleher@adaptcentre.ie
Technological University Dublin, Dublin, Ireland

**Abstract**

Preservation of domain knowledge from the source to target is crucial in any translation workflow. It is common in the translation industry to receive highly specialized projects, where there is hardly any parallel in-domain data. In such scenarios where there is insufficient in-domain data to fine-tune Machine Translation (MT) models, producing translations that are consistent with the relevant context is challenging. In this work, we propose a novel approach to domain adaptation leveraging state-of-the-art pretrained language models (LMs) for domain-specific data augmentation for MT, simulating the domain characteristics of either (a) a small bilingual dataset, or (b) the monolingual source text to be translated. Combining this idea with back-translation, we can generate huge amounts of synthetic bilingual in-domain data for both use cases. For our investigation, we use the state-of-the-art Transformer architecture. We employ mixed fine-tuning to train models that significantly improve translation of in-domain texts. More specifically, in both scenarios, our proposed methods achieve improvements of approximately 5-6 BLEU and 2-3 BLEU, respectively, on the Arabic-to-English and English-to-Arabic language pairs. Furthermore, the outcome of human evaluation corroborates the automatic evaluation results.

## 1 Introduction

Neural Machine Translation (NMT) has the ability to produce good quality translations in terms of both fluency and adequacy (Bahdanau et al., 2015). Nevertheless, NMT still faces some challenges when it comes to translation of out-of-domain texts (Koehn and Knowles, 2017). Domain adaptation of MT systems on in-domain parallel texts has been an active area of research to handle this situation. Among popular contributions to the domain adaptation research, Luong and Manning (2015) proposed to adapt an already existing NMT system to a new domain, with further training on the in-domain data only. In an effort to avoid overfitting on the in-domain data, Chu et al. (2017) employed the mixed fine-tuning approach, resuming training the baseline NMT model on a mix of in-domain and out-of-domain data. Other researchers suggested adding domain tags to either the source or target sentences of the in-domain data, to inform the NMT model about the domain during training and decoding (Britz et al., 2017; Kobus et al., 2017; Stergiadis et al., 2021).

In this sense, several research works on domain adaptation assume the availability of in-domain data. However, in-domain data scarcity is common in translation settings, due to the lack of specialized datasets and terminology, or inconsistency and inaccuracy of available in-domain translations. To tackle this problem, researchers have proposed diverse approaches, such as utilizing large monolingual datasets through selecting instances related to a given test set, then automatically translating this source-synthetic corpus, and finally fine-tuning the general NMT system on this data (Chinea-Ríos et al., 2017). Similarly, some works have investigated retrieving similar translations (fuzzy matches) from bilingual datasets, and then applying on-the-fly domain adaptation through fine-tuning the baseline model at translation time (Farajian et al., 2017), or integrating them into NMT training (Bulte and Tezcan, 2019; Xu et al., 2020).

While the aforementioned approaches prove to be helpful in certain scenarios of domain adaptation, we believe there is a need for further research in this area to address current challenges of in-domain data scarcity and synthetic data creation. Some approaches, such as on-the-fly domain adaptation, require using GPUs synchronously at translation time, which presents a challenge for some institutions due to the lack of resources. When it comes to mining monolingual or bilingual datasets for similar instances, in several domains a good similar sentence can be a mix of portions of multiple sentences. Besides, with the lack of in-house specialized translation memories, mining publicly available datasets can be an inefficient process.

In this work, we introduce a new approach to MT domain adaptation, leveraging state-of-the-art pre-trained language models (LMs) for domain-specific data augmentation. Our method can generate an unlimited number of in-domain sentences out of the box. Recently, there has been a considerable advancement in training large LMs (Radford et al., 2019; Brown et al., 2020; Black et al., 2022; Zhang et al., 2022), not only for English, but also for diverse languages (Antoun et al., 2021; Zhang et al., 2021; Müller and Laurent, 2022). More specifically, our current work exploits GPT-J (Wang and Komatsuzaki, 2021) and mGPT (Shliazhko et al., 2022) to generate texts from in-domain sentences. We investigate the feasibility of this domain-specific text generation technique when either no or limited bilingual in-domain dataset is available. Incorporating this approach in a process of bilingual in-domain synthetic data creation and then fine-tuning our baseline generic MT model on the new dataset (cf. Section 3), we report significant improvements of the translation quality of the in-domain test set (cf. Section 5).

The rest of the paper is organized as follows. In Section 2, we discuss the related work in detail. Then, we present our methods in Section 3. In Section 4, we describe the experimental setup and present the results of our experiments in Section 5. Finally, we conclude the paper and discuss future work in Section 6.

## 2 Related Work

In recent years, several pre-trained large LMs have been made available to the research community, covering a wide range of linguistic tasks. Among the state-of-the-art LMs are GPT-2 (Radford et al., 2019), BERT (Devlin et al., 2019), RoBERTa (Liu et al., 2019), XLNet (Yang et al., 2019), GPT-3 (Brown et al., 2020), ELECTRA (Clark et al., 2020), DeBERTa (He et al., 2020, 2021), T5 (Raffel et al., 2020), Gopher (Rae et al., 2021), GPT-J (Wang and Komatsuzaki, 2021), GPT-NeoX (Black et al., 2022), PaLM (Chowdhery et al., 2022), Chinchilla (Hoffmann et al., 2022), ELMFOREST (Li et al., 2022), MT-NLG (Smith et al., 2022), and OPT (Zhang et al., 2022). Some of these models are multilingual, such as BLOOM (BigScience, 2022), AlexaTM (FitzGerald et al., 2022) and mGPT (Shliazhko et al., 2022).

Using LMs for specialized domains has been explored by previous works for diverse tasks. Researchers explored the possibility to retrieve factual knowledge from LMs in various domains (Petroni et al., 2019; Sung et al., 2021). Similarly, Horawalavithana et al. (2022) developed large-scale models of foundational scientific knowledge that can be effectively used to perform a wide range of in-domain and out-of-domain tasks.

LMs have been used in Unsupervised NMT (Lample and Conneau, 2019; Chronopoulou et al., 2021; Wang et al., 2021). Large-scale pre-trained LMs have also been employed in a variety of MT tasks, to improve the robustness of MT models or their ability to work on domain texts (Bawden et al., 2020; Specia et al., 2020; Wenzek et al., 2021).

Recently, Chang et al. (2021) aimed at addressing the lack of training data for new application domains for data-to-text generation. They automatically augmented the data available for training by (a) generating new text samples by replacing specific values with alternative ones from the same category, (b) generating new text samples using GPT-2, and (c) proposing an automatic method for pairing the new text samples with data samples. Their approach boosted the performance of a standard seq2seq model by over 5 BLEU points. Sawai et al. (2021) investigated the use of GPT-2 for source-side data augmentation to improve the robustness of a generic pre-trained NMT model. They first fine-tuned the pre-trained model, BERT-fused (Zhu et al., 2020), on authentic bilingual data. Then, they augmented the English source with data generated by GPT-2. Thereafter, they forward-translated the source-side English monolingual data with the fine-tuned version of BERT-fused. Finally, they fine-tuned the model on a combination of the authentic and synthetic data. While the reported results showed reasonable improvement (approx. 2.0 BLEU points) for the English-to-Japanese language direction, insignificant improvement (avg. 0.3 BLEU) was achieved for both English-to-German and English-to-Chinese language directions. The authors concluded that the result could be due to the relatively small amount of the original English-to-Japanese data compared to the other two language directions. We conjecture that more factors might have led to this result, including using forward-translation (rather than back-translation) of a huge amount of data, due to the noise it introduces for the decoder (Haddow et al., 2022). In our current work, we try to be more specific about the task description, focussing on domain adaptation in the absence of enough in-domain data; utilizing back-translation as an effective data augmentation technique (Edunov et al., 2018; Caswell et al., 2019); and giving more attention to data distribution through applying approaches like mixed fine-tuning and oversampling (Chu et al., 2017).

Back-translation (Sennrich et al., 2016; Fadaee and Monz, 2018; Poncelas et al., 2019) corresponds to the scenario where target-side monolingual data is translated using an MT system to give corresponding synthetic source sentences, the idea being that it is particularly beneficial for the MT decoder to see well-formed sentences (Haddow et al., 2022). Back-translation has become a popular strategy among MT researchers, especially in low-resource scenarios (Haque et al., 2021). Burlot and Yvon (2018) performed a systematic study, which showed that forward-translation might lead to some improvements in translation quality, but not nearly as much as back-translation. Bogoychev and Sennrich (2019) concluded that forward-translation is more sensitive to the quality of the system used to produce synthetic data. Compared to back-translation, biases and errors in synthetic data are intuitively more problematic in forward-translation, since they directly affect the gold labels. The authors also reported that human evaluators favoured their back-translation systems over forward-translation systems, mostly in terms of fluency, while adequacy was largely the same across all of them, especially on the original translation direction. In their analysis, Edunov et al. (2018) showed that sampling or noisy synthetic data gives a much stronger training signal than data generated by beam or greedy search. Caswell et al. (2019) proposed a simpler alternative to noising techniques, consisting of tagging back-translated source sentences with an extra token. Hoang et al. (2018) empirically showed that the quality of the back-translation system matters for synthetic corpus creation, and that NMT performance can be improved by iterative back-translation in both high-resource and low-resource scenarios.

When it comes to fine-tuning strategies for MT domain adaptation, researchers demonstrated that applying the right data distribution can significantly mitigate catastrophic forgetting of strong baselines in domain adaptation settings. Chu et al. (2017) proposed the mixed fine-tuning method, whose training procedure is as follows: (a) train an NMT model on out-of-domain data until convergence, and (b) resume training the NMT model from the first step on a mix of in-domain and out-of-domain data (by oversampling the in-domain data) until convergence. According to the authors, mixed fine-tuning can address the overfitting problem of regular fine-tuning. In addition, mixed fine-tuning does not worsen the quality of out-of-domain translations, while regular fine-tuning does. Similarly, Hasler et al. (2021) studied the problem in an adaptation setting where the goal is to preserve the existing system quality while incorporating data for domains that were not the focus of the original MT system. They found that they could improve over the performance trade-off offered by Elastic Weight Consolidation (Kirkpatrick et al., 2017) with a relatively simple data mixing strategy.

# 3 Methods

In this work, we investigate two scenarios of in-domain data scarcity, and propose approaches to leverage pre-trained LMs for domain-specific data generation for MT training.

## 3.1 Use Case 1: Limited bilingual in-domain data available

This is a common scenario where a specialized translation project is received, and although there is a large bilingual generic dataset and a small bilingual in-domain dataset (e.g. translation memory), the in-domain data is insufficient for fine-tuning a baseline model. From now on, we will refer to this use case as "Setup 1". To handle this situation, we propose the following steps:

1. We employ text generation with a large LM in the target language to augment the in-domain data. In this process, each target sentence in the in-domain dataset is used as a prompt to generate synthetic segments using the pre-trained language model. As expected, the generated text preserves the domain characteristics of the authentic in-domain data. This step enables us to have sufficient data in the target language.

2. To obtain parallel source sentences, we back-translate the target-side synthetic sentences that were generated in the previous step.

3. We apply mixed fine-tuning proposed by Chu et al. (2017) to the baseline model. In other words, we continue training our baseline model on a mix of (a) the synthetic bilingual in-domain dataset we got from the two previous steps, and (b) a randomly sampled portion of the original generic dataset, with a data size ratio of 1:9, respectively. To apply oversampling, we employ the dataset weights feature in OpenNMT-tf[1] (Klein et al., 2020), with weights 0.9 and 0.1, respectively. Hence, the dataset weights are inversely proportional to the sizes of the two datasets.[2] As the in-domain corpus is smaller than the generic corpus, oversampling allows the model to pay equal attention to both corpora. As a result of the mixed fine-tuning process, we obtained a new model that translates in-domain data significantly better than the baseline (cf. Section 5).[3]

4. Although the new fine-tuned model can still adequately translate generic data, we noticed it can degrade performance by 1-2 BLEU points. Therefore, we experimented with checkpoint averaging (Vaswani et al., 2017) of the fine-tuned model with the baseline model to reduce variability between trainings and address rapid overfitting during fine-tuning (Tran et al., 2021). This step helps regain the higher evaluation score of the baseline model on generic data, while retaining the improved score of the fine-tuned model on in-domain data.

## 3.2 Use Case 2: Zero bilingual in-domain data available

In this case, we assume that there is no bilingual in-domain data at all. There is only the source text that requires translation. From now on, we will refer to this use case as "Setup 2".

The first step is to use the baseline MT model for forward-translation of the source text. The generated translation might not be perfect; however, it can still include useful information about the domain. This approach bootstraps some parallel data for a situation where there was none. Then, we follow the same four steps mentioned in the previous use case.

---

[1] https://github.com/OpenNMT/OpenNMT-tf

[2] This configuration creates a weighted dataset where examples are randomly sampled from the data files according to the provided weights. In simple words, it sequentially samples 9 examples from the smaller in-domain dataset, and 1 example from the larger generic dataset, and so on.

[3] Inspired by Hasler et al. (2021) who applied 20x oversampling, we experimented with a higher oversampling ratio. Increasing both the data size and weight degraded performance on the in-domain test set, compared to our applied 9x ratio, while increasing the weight only did not result in a significant improvement. We might investigate the effect of changing the oversampling ratio further in the future.

## 4 Experiment Setup

### 4.1 Datasets

For training Arabic-to-English and English-to-Arabic generic models, we collect high-quality datasets from OPUS (Tiedemann, 2012). The breakdown of segment numbers in our datasets before and after filtering is shown in Table 1. To ensure the quality of our datasets, we apply a multi-filtering process. First, we apply rule-based filtering to individual datasets, removing duplicates, source-copied segments, those with too long source/target (ratio 200% and > 200 words), and HTML tags. Then, we calculate the similarity between each source and target to semantically filter out segments with a similarity threshold lower than 0.45. Finally, we concatenate the datasets and apply global filtering. For the development and test datasets, we randomly sampled 5000 segments each from the original dataset.[4]

For in-domain NMT models, we use TICO-19 (Anastasopoulos et al., 2020), a dataset in the Public Health domain. After filtering, the dataset includes 3062 segments. Table 2 shows the dataset details. We split the TICO-19 dataset into a development dataset, with 1000 segments, and a test dataset which includes the rest, i.e. 2062 segments. The whole TICO-19 dataset is used for generating a large synthetic in-domain training dataset, as described in Section 4.5.

|                  |            | Filtering   |            |
| ---------------- | ---------- | ----------- | ---------- |
| **Dataset**      | **Raw**    | **Rule-based** | **Semantic** |
| Bible            | 62,195     | 47,699      | 43,951     |
| ELRC_2922        | 15,129     | 14,937      | 14,850     |
| GlobalVoices     | 63,071     | 55,201      | 51,220     |
| GNOME            | 150        | 143         | 134        |
| Infopankki       | 50,769     | 15,531      | 14,635     |
| KDE4             | 116,239    | 85,003      | 68,180     |
| MultiUN          | 9,759,125  | 7,807,811   | 7,508,443  |
| News-Commentary  | 97,384     | 80,744      | 77,715     |
| OpenSubtitles    | 29,823,188 | 23,666,245  | 20,176,228 |
| Tatoeba          | 27,905     | 27,649      | 26,714     |
| Ubuntu           | 5,978      | 5,617       | 5,340      |
| UN               | 74,067     | 63,074      | 62,901     |
| UNPC             | 20,044,478 | 15,696,210  | 15,441,996 |
| Wikimedia        | 407,543    | 335,783     | 317,285    |
| Wikipedia        | 151,136    | 117,859     | 116,940    |
| **Total**        | **60,698,357** | **48,019,506** | **43,926,532** |
| **Global Filtering** |        | **40,207,905** |          |

Table 1: Generic datasets

|         |       | Filtering   |          |
| ------- | ----- | ----------- | -------- |
| **Dataset** | **Raw** | **Rule-based** | **Semantic** |
| TICO-19 | 3,071 | 3,069       | 3,062    |

Table 2: In-domain dataset (Public Health)

### 4.2 Vocabulary

To create our vocabulary, we first train SentencePiece unigram models (Kudo and Richardson, 2018; Kudo, 2018) for the source and target individually, to learn subword units from unto-kenized text.[5] Then, we utilize this SentencePiece model to subword our dataset. We use a vocabulary size of 50,000. Subsequently, we convert the learned subword units into our final vocabulary in the format supported by OpenNMT-tf. Segments are automatically augmented with start and end tokens via *source_sequence_controls* option.

---

[4]Our MT preparation scripts are publicly available at: https://github.com/ymoslem/MT-Preparation

### 4.3 NMT Model Architecture

Our baseline generic NMT models use the Transformer "Big" architecture (Vaswani et al., 2017) as implemented in OpenNMT-tf, and relative position representations (Shaw et al., 2018) with a clipping distance k=20. The models consist of 6 layers with a model dimension of 1,024, split into 16 heads, and a feedforward dimension of 4,096.

### 4.4 Training

The training takes place on 2x NVIDIA RTX A4000 GPUs, with a batch size of 2048 tokens per GPU, for an effective batch size of 25k tokens/step. The Arabic-to-English model is trained for 240k steps, while the English-to-Arabic model is trained for 105k steps. Early stopping is used after 3 evaluations with less than 0.01 BLEU improvement on the development dataset.

### 4.5 Domain-Specific Data Generation with LMs

For English, we use GPT-J (Wang and Komatsuzaki, 2021), a Transformer-based language model with 6B trainable parameters.[6] For Arabic, we use mGPT (Shliazhko et al., 2022), a multilingual language model.[7]

To fit the models onto an NVIDIA RTX A4000 GPU (16 GB of GPU memory), the half-precision floating-point (float16) format is used.[8] We also use a batch size of 1.[9] For inference, we employ 50 Top-K sampling and 0.95 Top-p (nucleus) sampling (Fan et al., 2018; Holtzman et al., 2018; Radford et al., 2019; Holtzman et al., 2020). The maximum length of the generated text is set to 300 tokens, and we return 5 sequences for each segment, to get multiple independently sampled outputs. Finally, we split the generated text into sentences.[10]

As explained in Section 3, we have two use cases: (a) a small bilingual in-domain dataset is available; and (b) the source only is available, so we utilize forward-translation to generate the target side. After that, each target sentence of the in-domain dataset TICO-19 (i.e. the authentic target in the first case, or the MT-ed target in the second case) is fed to the LM as a prompt to generate synthetic in-domain segments. We use random seeds to generate multiple datasets, namely 2 for English and 3 for Arabic.[11] We filter the concatenated datasets, by removing duplicates and cleaning lines with a wrong language, and those including only dashes or filenames. Table 3 illustrates the numbers of in-domain synthetic segments generated by the LMs.

| Language | LM | Setup 1 | | | | | Setup 2 | | | | |
|---|---|---|---|---|---|---|---|---|---|---|---|
| | | 1st Run | 2nd Run | 3rd Run | Total | Filtered | 1st Run | 2nd Run | 3rd Run | Total | Filtered |
| **English** | GPT-J | 131,730 | 131,554 | N/A | 263,284 | 242,469 | 137,705 | 138,702 | N/A | 276,407 | 253,287 |
| **Arabic** | mGPT | 96,296 | 97,031 | 94,513 | 287,840 | 271,665 | 103,272 | 103,459 | 103,303 | 310,034 | 294,391 |

Table 3: Data generated by language models (LMs)

---

[5]In SentencePiece, we utilize the training options `--split_digits` to split all digits into separate pieces, and `--byte_fallback` to decompose unknown pieces into UTF-8 byte pieces to help avoid out-of-vocabulary tokens.

[6]https://huggingface.co/EleutherAI/gpt-j-6B

[7]https://huggingface.co/sberbank-ai/mGPT

[8]In Hugging Face Transformers, we also set the option `low_cpu_mem_usage` to `True`.

[9]It is worth mentioning though that for batch generation (i.e. >1), padding and attention masking should be used; note that left padding is required for GPT-like models.

[10]Our scripts are available at: https://github.com/ymoslem/MT-LM

[11]As two data generation runs for Arabic resulted in a less amount of data than for English, we increased the data size for Arabic by generating a third dataset (cf. Table 3).

### 4.6 Back-Translation

For back-translation, we use OPUS models,[12] specifically the Transformer-Big versions. For efficiency purposes, we convert the models to the CTranslate2[13] format (INT8 quantization). We use beam size 5. After back-translation, we run the same rule-based and semantic filtering on the generated dataset as we did for the original datasets. Table 4 elaborates on the numbers.

| Language | Setup 1 | | | Setup 2 | | |
|---|---|---|---|---|---|---|
| | Translated | Filtering | | Translated | Filtering | |
| | | Rule-based | Semantic | | Rule-based | Semantic |
| **English** | 242,469 | 240,329 | 239,931 | 253,287 | 251,357 | 250,317 |
| **Arabic** | 271,665 | 271,645 | 270,743 | 294,391 | 294,234 | 293,252 |

Table 4: Back-translated datasets

### 4.7 Mixed Fine-tuning

Following Chu et al. (2017), we employ the mixed fine-tuning approach. We randomly sample a portion from the generic data we used to train the baseline model, and use it during the fine-tuning step along with the in-domain dataset. Oversampling the in-domain data is a crucial step, as explained in Section 3. We first train a baseline NMT model on out-of-domain data until convergence, and then continue training the NMT baseline model on a mix of in-domain and out-of-domain data (by oversampling the in-domain data) until convergence.

In most experiments, we fine-tuned the baseline for 5000 steps. However, for Setup 2 of the English-to-Arabic language pair, we found that the best automatic evaluation scores were achieved with training for only 500 or 1000 steps. We believe that this might be due to the quality or distribution of the generated in-domain data compared to the original generic data. Although Chu et al. (2017) observed that both regular fine-tuning and mixed fine-tuning tend to converge after 1 epoch of training, it seems there is no golden rule as to how many steps or epochs the baseline model should be fine-tuned on the mixed data. Depending on the size of data, we recommend conducting less-frequent evaluations on the development dataset during the fine-tuning process for finding out the best model checkpoint.

## 5 Results

In this section, we elaborate on our automatic and human evaluations and discuss the results. As Table 5 shows, scores obtained from diverse automatic metics provide good correlation with the human evaluation. Moreover, the linguistic analysis (cf. Section 5.3) supports these numerical results, and demonstrates how the models fine-tuned on synthetic in-domain data produce more accurate translations of the in-domain test set compared to the baseline model.

### 5.1 Automatic Evaluation

For automatic evaluation, we calculated spBLEU (Papineni et al., 2002; Goyal et al., 2022) which uses a SentencePiece tokenizer with 256,000 tokens and then the BLEU score is computed on the sub-worded text. spBLEU has been recently added to sacreBLEU v2.1.0.[14] Goyal et al. (2022) showed that spBLEU exhibits a strong correlation with the tokenization-independent chrF++, yet has the advantage of keeping the familiarity of BLEU. To verify our results, we employed other evaluation metrics, namely the character-based metric chrF++ (Popović, 2017), and the word-based metric TER (Snover et al., 2006), as implemented in sacre-BLEU (Post, 2018). Furthermore, we integrated COMET[15] (Rei et al., 2020) as a semantic

---
[12] https://github.com/Helsinki-NLP/Tatoeba-Challenge/tree/master/models
[13] https://github.com/OpenNMT/CTranslate2
[14] https://github.com/mjpost/sacrebleu
[15] https://github.com/Unbabel/COMET

evaluation metric, with the "wmt20-comet-da" model.

We experimented with averaging parameters across multiple model checkpoints (Vaswani et al., 2017), to address bias towards recent training data (Tran et al., 2021). Sometimes, averaging multiple checkpoints of a baseline model, or averaging a baseline model with a fine-tuned model could lead to extra improvements of the automatic and/or human evaluation of our models. Table 5 shows evaluation results on the in-domain test dataset, and Figure 1 elaborates on all the automatic evaluation results, including the results for averaged models.

| Language | Model | spBLEU ↑ | chrF++ ↑ | TER ↓ | COMET ↑ | Human ↑ |
|---|---|---|---|---|---|---|
| **AR-EN** | Baseline | 44.57 | 66.68 | 46.67 | 65.78 | 87.0 |
| | Setup 1 Mixed Fine-Tuning | 49.79 | 70.54 | 43.32 | 71.89 | 93.5 |
| | Setup 2 Mixed Fine-Tuning | 47.22 | 69.38 | 45.38 | 70.08 | 94.5 |
| **EN-AR** | Baseline | 36.15 | 58.3 | 58.29 | 57.5 | 87.0 |
| | Setup 1 Mixed Fine-Tuning | 42.38 | 62.52 | 53.99 | 67.48 | 90.0 |
| | Setup 2 Mixed Fine-Tuning | 37.91 | 59.42 | 55.95 | 59.47 | 88.5 |

Table 5: Evaluation results on the in-domain test set, TICO-19

## 5.2 Human Evaluation

Since translation focusses mainly on word choice, syntax, and semantics, and how people perceive it, we decided to complement our evaluation process with human evaluation.

The evaluator was an Arabic native speaker and domain expert. We conducted a bilingual evaluation, providing the evaluator with both the original source sentences and translations generated by the MT models. The human test set contained 50 sentences, randomly extracted from the original test set, and verified as accepted translations. The evaluator was asked to assess the acceptability of each translation generated by our baselines and fine-tuned MT systems, using the scale proposed by Coughlin (2003), ranging from 1 to 4, and outlined as follows:

- **4 = Ideal:** Not necessarily a perfect translation, but grammatically correct, with all information accurately transferred.
- **3 = Acceptable:** Not perfect (stylistically or grammatically odd), but definitely comprehensible, AND with accurate transfer of all important information.
- **2 = Possibly Acceptable:** Possibly comprehensible (given enough context and/or time to work it out); some information transferred accurately.
- **1 = Unacceptable:** Absolutely not comprehensible and/or little or no information is accurately transferred.

Human evaluation results on the in-domain dataset, TICO-19, are expressed in percentage points in the last column of Table 5. In addition, Table 6 elaborates on the results for all the systems, showing the mean value for each system on the 1-4 scale.[16] The models fine-tuned on the domain-specific synthetic dataset achieve improvements on the in-domain test set, while retaining the baseline's quality on the generic holdout test set.

| Language | Test Set | BS | BS-Avg8 | MixFT-1 | MixFT-1+BS | MixFT-1+BS-Avg8 | MixFT-2 | MixFT-2+BS | MixFT-2+BS-Avg8 |
|---|---|---|---|---|---|---|---|---|---|
| **AR-EN** | Generic | 3.84 | **3.90** | 3.84 | 3.88 | 3.88 | 3.84 | 3.84 | 3.84 |
| | TICO-19 | 3.48 | 3.62 | 3.74 | **3.82** | 3.80 | 3.78 | 3.72 | 3.74 |
| **EN-AR** | Generic | **3.96** | 3.90 | 3.82 | **3.96** | 3.90 | 3.94 | **3.96** | **3.96** |
| | TICO-19 | 3.48 | 3.50 | **3.60** | 3.54 | 3.52 | 3.54 | 3.56 | 3.54 |

Table 6: Human evaluation of MT models for Arabic-to-English (AR-EN) and English-to-Arabic (EN-AR) language pairs, the baseline (BS), baseline averaged 8 checkpoints (BS-Avg8), mixed fine-tuning model (MixFT), averaging MixFT with BS (MixFT+BS), and averaging MixFT with BS-Avg8 (MixFT+BS-Avg8). MixFT-1 refers to Setup 1 and MixFT-2 refers to Setup 2.

---

[16]Sentence-level human evaluation is available at: `https://github.com/ymoslem/MT-LM`

## 5.3 Linguistic Analysis

We observe that in several cases, the fine-tuned (in-domain) models generate more idiomatic translations or better capture the meaning in the Public Health context. Samples from the test dataset translated by the baseline model and in-domain models reflect these improvements.

Among Arabic-to-English examples, the phrase "غير مسببة للأمراض في مضيفاتها المستودعة الطبيعية" was translated as "not pathogenic in their naturally occurring host" by the baseline, and "non-pathogenic in their natural reservoir hosts" by both in-domain models. The former translation somehow conveys the meaning; however, the latter translation is more idiomatic in the medical context. The baseline system translated "حمامات الولادة" as "maternity wards" which is an incorrect translation, while the in-domain models in Setup 1 and Setup 2 produced more relevant translations as "birthing pools" and "birth baths", respectively. The baseline model translated "مسحة بلعومية أنفية" as "a nasal laryngeal swab" which is an inaccurate translation. In contrast, both in-domain models translated the term as "a nasal nasopharyngeal swab", which uses the accurate "nasopharyngeal" medical term. It can still be edited by removing the redundant "nasal"; however, our evaluator gave it a higher score than the translation provided by the baseline. The term "اختبارات مَصلِيَّة" was translated as "serum tests" by the baseline, while it was translated as "serological tests" by both in-domain models, which is more idiomatic.

Examining some of the English-to-Arabic translations, the baseline model mistranslated "HCoVs" as "فيروسات نقص المناعة البشرية / متلازمة نقص المناعة المكتسب (الإيدز)" (HIV/AIDS), as opposed to the in-domain models, which correctly translated it as "فيروسات كورونا البشرية" or just "HCoV فيروسات". Interestingly, even for a simpler phrase like "five times more cases", the baseline incorrectly translated it as "خمس حالات" (five cases), whilst the in-domain models correctly conveyed the meaning as "خمسة أضعاف الحالات".

There are also examples where one of the in-domain systems generated the correct translation while the other could not. For instance, both the baseline and Setup 2 in-domain model translated "If you do wear a mask" as "إذا كنت لا ارتداء قناع", which is both syntactically and semantically incorrect. In contrast, the Setup 1 in-domain model perfectly translated it as "إذا كنت ترتدي قناعًا". The baseline model translated the phrase "passive antibody therapy" as "العلاج السلبي للأجسام المضادة", which uses the preposition "لـ" (of) instead of "بـ" (with), missing the fact that in this context "antibody" is equivalent to "antibody-based" rather than being the issue to be treated. Similarly, the Setup 2 in-domain model mistranslated it as "العلاج المضاد السلبي" while the Setup 1 in-domain model accurately translated it as "العلاج السلبي بالأجسام المضادة".

Since some phrases can be expressed in multiple ways, we notice that sometimes the evaluator equally ranked different translations. This might explain why automatic metrics evaluate Arabic-to-English Setup 1 higher than Setup 2, whereas the human evaluation shows that the translation quality of both setups is comparable.

## 6 Conclusion

In this paper, we propose two simple methods to utilize pre-trained language models for domain-specific data augmentation for NMT systems. We report significant improvements, supported by both automatic and human evaluation. The proposed techniques enable the generation of large amounts of data, simulating the characteristics of the specialized text to be translated, and facilitating the domain adaptation process.

For the Arabic-to-English language direction, human evaluation demonstrates that Setup 2 is on par with Setup 1 even though in the former we did not have any authentic bilingual in-domain data (cf. Section 3). Nevertheless, the English-to-Arabic model in Setup 2 has lower performance compared to the Setup 1 model, although both setups outperform the baseline on the in-domain test set. We believe this might be due to the quality of synthetic data generated for Arabic, which is an interesting aspect to explore further.

In the future, we would like to investigate utilizing terminology for domain-specific data generation, and experiment with employing the same proposed approaches for low-resource languages and multilingual settings.

**Figures: Automatic Evaluation**

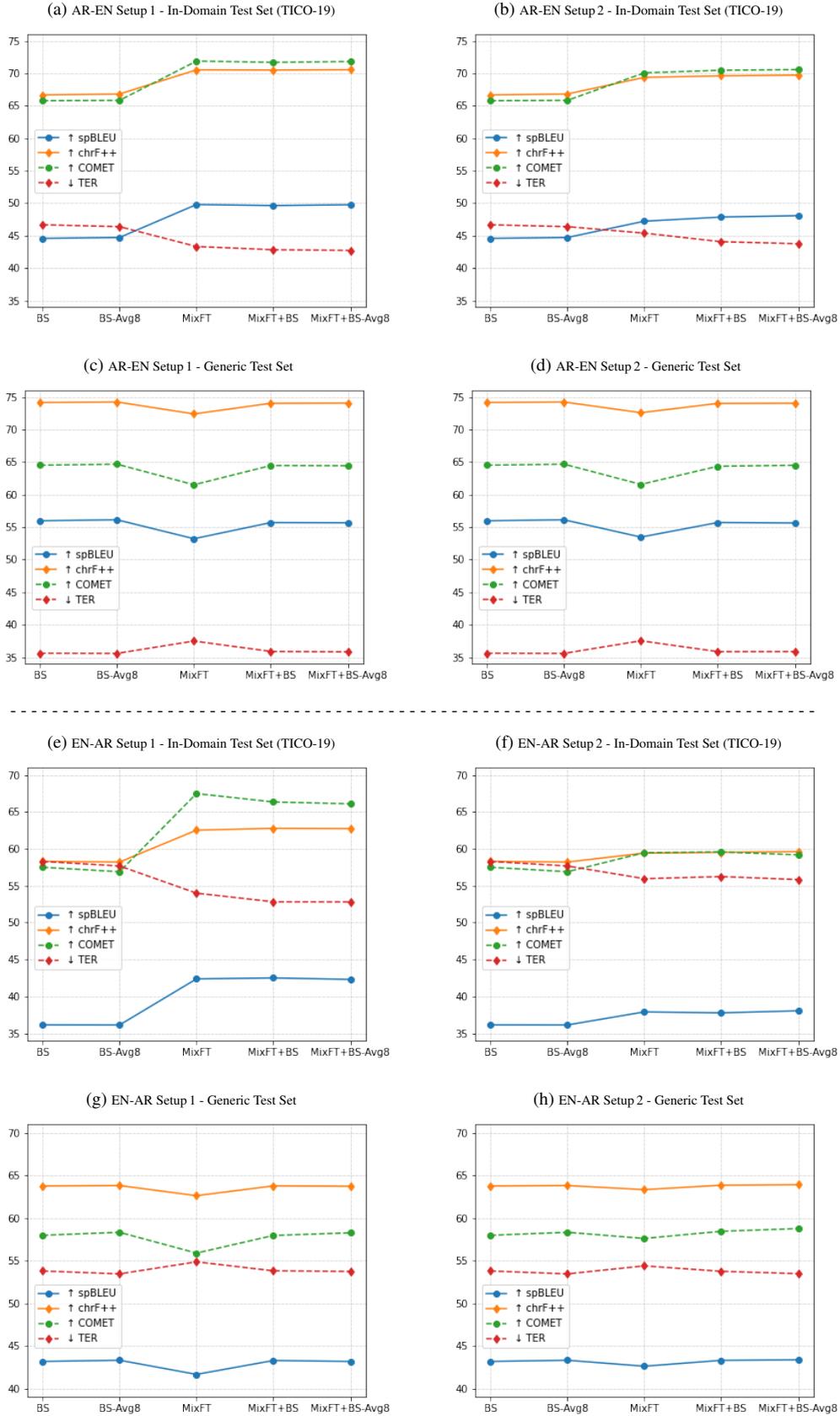

Figure 1: Performance comparison of 5 models for Arabic-to-English (AR-EN) and English-to-Arabic (EN-AR) language pairs, the baseline (BS), baseline averaged 8 checkpoints (BS-Avg8), mixed fine-tuning model (MixFT), averaging MixFT with BS (MixFT+BS), and averaging MixFT with BS-Avg8 (MixFT+BS-Avg8). The MixFT models fine-tuned on the domain-specific synthetic dataset achieve improvements on the in-domain test set (a,b & e,f), while retaining the baselines quality on the generic test set (c,d & g,h).

# 7 Acknowledgements

This work is supported by the Science Foundation Ireland Centre for Research Training in Digitally-Enhanced Reality (d-real) under Grant No. 18/CRT/6224, the ADAPT Centre for Digital Content Technology which is funded under the Science Foundation Ireland (SFI) Research Centres Programme (Grant No. 13/RC/2106) and is co-funded under the European Regional Development Fund, and Microsoft Research.

We would like to extend our sincere thanks to Dr Muhammed Yaman Muhaisen, Ophthalmologist at Eye Surgical Hospital (Damascus), for conducting the human evaluation of our machine translation models.